\documentclass[runningheads]{llncs}
\usepackage{graphicx}
\usepackage{amsmath}
\usepackage{amsfonts}
\usepackage{multirow}
\usepackage{float}
\usepackage{multirow}
\usepackage{color}
\usepackage[colorlinks]{hyperref}
\newcommand{\bheading}[1]{{\noindent{\textbf{#1}}}}

\newcommand{\tabincell}[2]{\begin{tabular}{@{}#1@{}}#2\end{tabular}}

\begin{document}
\title{One-Shot Medical Landmark Detection}
%
%
\author{Qingsong Yao\inst{1, 2} \and
Quan Quan\inst{1, 2} \and
Li Xiao\inst{1} \and
S. Kevin Zhou\inst{1, 2}
}

\authorrunning{Q. Yao, et al.}

\institute{\
Medical Imaging, Robotics, Analytic Computing Laboratory/Engineering (MIRACLE), Key Lab of Intelligent Information Processing of Chinese Academy of Sciences (CAS), Institute of Computing Technology, CAS, Beijing 100190, China\\
 \email{\{yaoqingsong19\}@mails.ucas.edu.cn} \email{\{quanquan, xiaoli,zhoushaohua\}@ict.ac.cn}  \and
Peng Cheng Laboratory, Shenzhen, China}
\maketitle       
\begin{abstract}
The success of deep learning methods relies on the availability of a large number of datasets with annotations; however, curating such datasets is burdensome, especially for medical images. To relieve such a burden for a landmark detection task, we explore the feasibility of using \textbf{only a single annotated image} and propose a novel framework named Cascade Comparing to Detect (CC2D) for one-shot landmark detection. CC2D consists of two stages: 1)  Self-supervised learning (CC2D-SSL) and 2) Training with pseudo-labels (CC2D-TPL). CC2D-SSL captures the consistent anatomical information in a coarse-to-fine fashion by comparing the cascade feature representations and generates predictions on the training set. CC2D-TPL further improves the performance by training a new landmark detector with those predictions. The effectiveness of CC2D is evaluated on a  widely-used public dataset of cephalometric landmark detection, which achieves a competitive detection accuracy of 81.01\% within 4.0mm, comparable to the state-of-the-art fully-supervised methods using a lot more than one training image.

\keywords{Medical Landmark Detection \and One-shot Learning}
\end{abstract}
\section{Introduction}

Accurate and reliable anatomical landmark detection is a fundamental first step in therapy planning and intervention, thus it has attracted great interest from academia and industry~\cite{yang2017automatic,yao2020miss,zhou2019handbook,zhou2021review}. It has been proved crucial in many medical clinical scenarios such as knee joint surgery~\cite{yang2015automated}, bone age estimation~\cite{gertych2007bone}, carotid artery bifurcation~\cite{zheng20153d}, orthognathic and maxillofacial surgeries~\cite{chen2019cephalometric}, and pelvic trauma surgery~\cite{bier2018x}. Furthermore, it plays an important role in medical image analysis~\cite{liu2010search,payer2016regressing}, e.g.,  initialization of registration or segmentation algorithms.

Recently, deep learning based methods have been developed to efficiently localize anatomical landmarks in radiological images such as computed tomography (CT) and X-rays~\cite{bhalodia2020self,lang2020automatic,payer2016regressing,Wei2020Measurements}. Zhong et al.~\cite{zhong2019attention} use cascade U-Net to launch a two-stage heatmap regression. Chen et al.~\cite{chen2019cephalometric} regress the heatmap and coordinate offset maps at the same time. Liu et al.~\cite{Wei2020Measurements} improve the performance by utilizing relative position constraints. Li et al.~\cite{li2020gcn} adapt cascade deep adaptive graphs to capture relationships among landmarks.

\begin{figure}[t]
\centering
\includegraphics[width=0.8\textwidth]{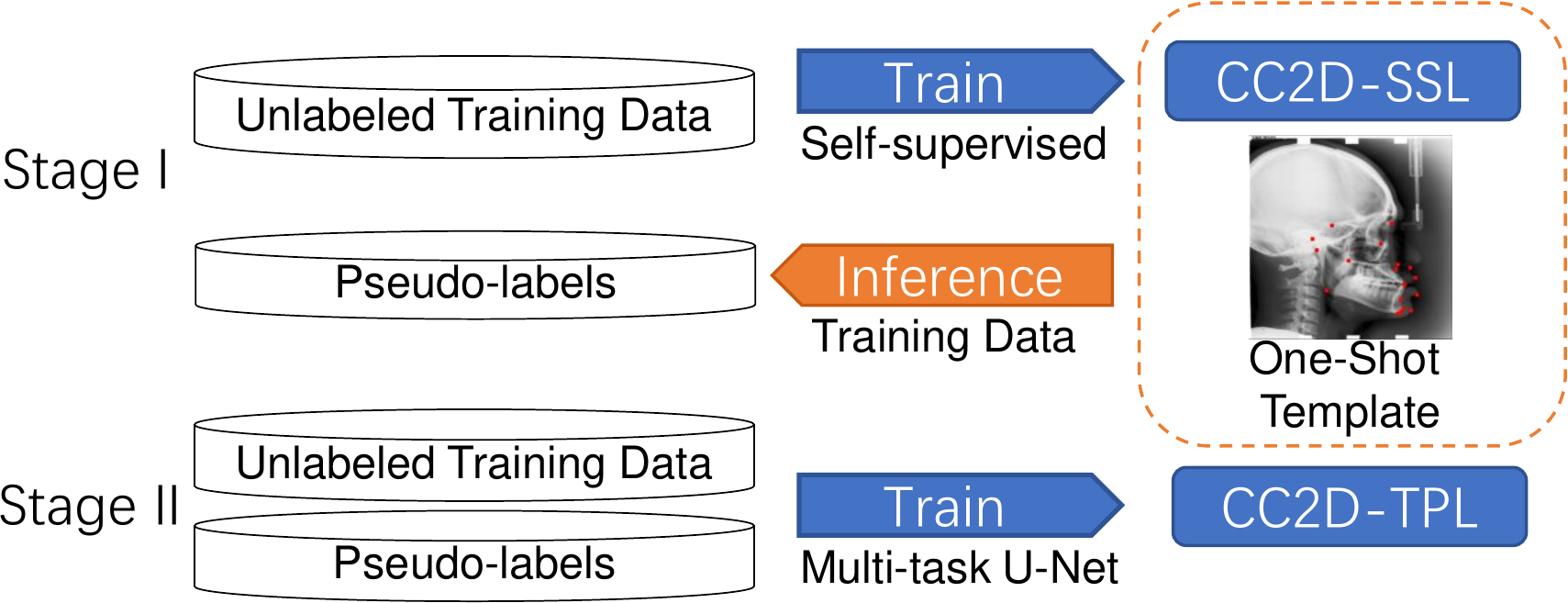}
\caption{Overview of the proposed Cascade Comparing to Detect (CC2D) framework. CC2D-SSL and CC2D-TPL represents self-supervised learning and training with pseudo-labels, respectively.} 
\label{fig:CC2D}
\end{figure}

As is well known, expert-annotated training data are needed for all of those supervised methods. A common wisdom is that a model with a better generalization is learned from more data. However, it often consumes considerable cost and time for the radiologists to annotate sufficient training data, which might restrict further application in clinical scenarios. Several self-supervised learning attempts have been explored to break this limitation in classification and segmentation tasks, including patch ordering~\cite{zhu2020rubik}, image restoration~\cite{zhou2019models,zhou2021models}, superpixel-wise~\cite{ouyang2020self} and patch-wise~\cite{chaitanya2020contrastive} contrastive learning. Nevertheless, training a landmark detection model with few annotated images is still challenging.

The landmarks are usually associated with anatomical features (e.g. the lips are halfway between the chin and nose in a cephalometric x-ray), and hence those features exhibit certain invariance among different images. Therefore, we pose an interesting question: {Can we determine landmarks by learning from a few or even just one labeled image?} In this paper, we challenge the hardest scenario: \emph{Only one annotated medical image is available}, which defines the number of the landmarks and their corresponding locations of interests. 

To tackle the challenge, we propose a novel framework named ``Cascade Comparing to Detect \textbf{(CC2D)}" for one-shot landmark detection. CC2D consists of two stages: 1) Self-supervised learning \textbf{(CC2D-SSL)} and 2) Training with pseudo-labels \textbf{(CC2D-TPL)}. 
The CC2D-SSL idea is motivated by contrastive learning~\cite{chen2020simple,zhou2020comparing}, which learns effective features for image classification. We instead propose to learn feature representations that embed consistent anatomical information for \textit{image patch matching}. Our design is further inspired by our observation learned through interactions with clinicians that they firstly roughly locate the target regions through a coarse screening and then progressively refine the fine-grained location. Therefore, we propose to match the coarse-grained corresponding areas first, then gradually compare the finer-grained areas in the selected coarse area. Finally, the targeted landmark is localized precisely by comparing the cascade embeddings in a coarse-to-fine fashion. 

In the second CC2D-TPL stage, we first use the CC2D-SSL model to generate predictions for the whole training set as \textit{pseudo-labels}, followed by training a CNN-based landmark detector using these pseudo-labels. This step brings two benefits. On one hand, the inference procedure becomes more concise. On the other hand, recent findings show that training an over-parameterized network from scratch tends to learn noiseless information firstly~\cite{arpit2017closer}. In our case, as we cannot predict every training point as accurate as ground truth in the SSL stage, a newly trained landmark detector can improve the performance by capturing the regular information hidden from the noisy labels produced by the CC2D-SSL stage. 

We evaluate the performance of CC2D on the public dataset from the ISBI 2015 Grand-Challenge. CC2D achieves competitive detection accuracy of 81.01\% within 4.0mm, comparable to the state-of-the-art fully-supervised methods that use a lot more than one training image.

\begin{figure}[t]
\centering
\includegraphics[width=0.9\textwidth]{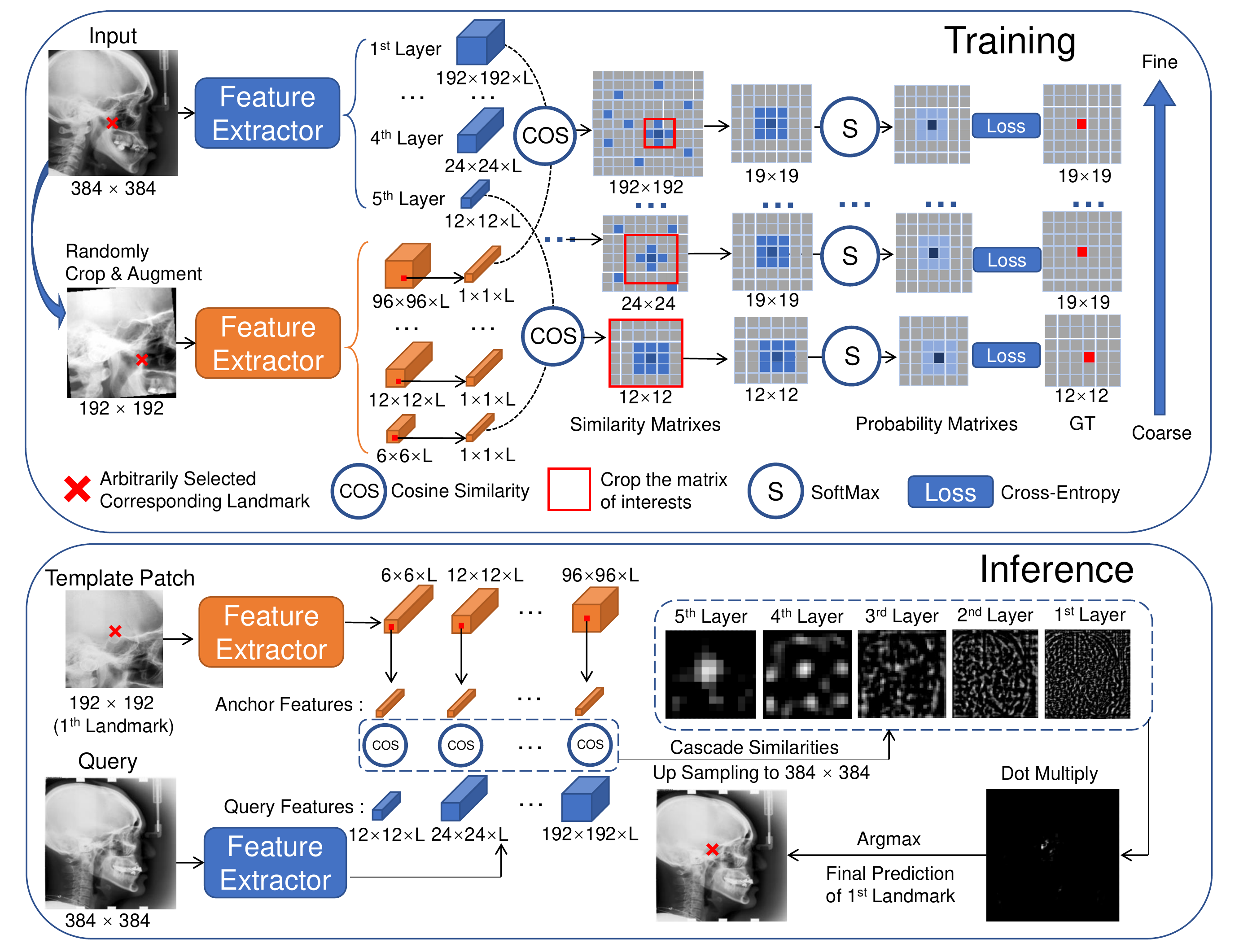}
\caption{Overview of the self-supervised learning part (CC2D-SSL). The two feature extractors are penalized to embed the corresponding landmarks to similar embeddings in the cascade feature space. The embeddings in the deepest layer select the coarse-grained area, while the embeddings in the inner layer improve the localization accuracy by comparing the finer areas in the selected coarse area. As consequence, the missing landmark is localized precisely in a coarse-to-fine fashion.} 
\label{fig:CC2D-SSL}
\end{figure}

\section{Method}

In this section, we first introduce the mathematical details of the training and inference stage of the self-supervised learning part of CC2D in Section~\ref{Sec:stage_1}, respectively. Then, we illustrate how to train a new landmark detector from scratch with pseudo-labels in Section~\ref{Sec:Stage_2}, which are the predictions of CC2D-SSL on the training set. The resulting detector is used to predict results for the test set.


\subsection{Stage I: Self-supervised Learning (CC2D-SSL)}
\label{Sec:stage_1}

\textbf{Training stage:} As shown in Fig.~\ref{fig:CC2D-SSL}, for an input image $X_r$ resized to $384\times384$, we arbitrarily select a target point $P_r=(x_r, y_r)$ and randomly crop a patch $X_p$ with size $192\times192$ which contains $P$. Then we apply data augmentation to the content of $X_p$ by rotation and color jittering. The chosen landmark is moved to $P_p=(x_p, y_p)$. Two feature extractors $E_r$ and $E_p$ project the input $X_r$ and the patch $X_p$ into a multi-scale feature space, resulting in a cascade of embeddings, denoted by $F_r$ and $F_p$ with a length of $L$, respectively. Here we mark the embedding of $i^{th}$ layer as $F^i$. Next, we extract the feature of the anchor point, denoted by $F_a^{i}$, from $F_p^{i}$ for each layer, guided by its corresponding coordinates $P_a^{i}=(x_p / 2^i, y_p / 2^i)$ which are down-sampled for $i$ times. To compare the features of input $F_r$ with anchor features $F_a$, we compute cosine similarity maps for each layer: 

\begin{equation}
\begin{aligned}
s^i =  \frac{\langle F_a^i \cdot F_r^i \rangle}{||F_a^i||_2 \cdot ||F_r^i||_2},
\end{aligned}
\label{Eq:cos}
\end{equation}

In CC2D-SSL, similarity maps at different scales have different roles. The deepest one ($s^5$ in this paper) differentiates the target coarse-grained area from every different area in the whole map, while the most shallow ones are only in charge of distinguishing the target pixel with adjacent pixels. The cascade similarity maps locate the target point in a \textbf{coarse-to-fine fashion}. Therefore, we set the matrix of interests $m^i=s^i$ if the $i^{th}$ layer is deepest. Otherwise, we crop the similarity map $s^i$ to a square patch $m^i$ centered on $P_a^{i}$ of size $\alpha$. As we aim at increasing the similarity of the correct pixels while decreasing others, we use softmax function to normalize $m^i$ to probability matrix $y^i$ with a temperature $\tau$: $y^i = softmax(m^i * \tau)$. Note that the softmax operation is performed on all pixels. Correspondingly, we set the ground truth $GT^i$ of each layer:
\begin{equation}
\begin{aligned}
GT^i =    \{  \begin{tabular}{l}
        $[\pi(x,y|x_p / 2^i, y_p / 2^i)]_{12 \times 12}$ ~~\text{If layer $i$ is the deepest};\\
      $[\pi(x,y|\alpha/2,\alpha/2)]_{\alpha \times \alpha}$ ~~\text{\qquad Otherwise},
  \end{tabular}
\end{aligned}
\label{Eq:GT}
\end{equation}
where $[\pi(x,y)]_{M \times N}$ is a matrix of size $M \times N$ and $\pi(x,y|x_0,y_0)$ is an indicator function that outputs 1 only if $x=x_0~\&~y=y_0$ and 0 otherwise.
We use cross-entropy loss $L_{CE}$ to penalize the divergence between probability matrix $y^i$ and ground-truth $GT^i$ for each layer and summarize the losses as final loss $L_{SSL}$:

\begin{equation}
\begin{aligned}
L_{SSL} =  \sum_i L_{CE}(y^i, GT^i),
\end{aligned}
\label{Eq:L_SSL}
\end{equation}



\textbf{Inference stage to generate pseudo-labels:} As shown in Fig.~\ref{fig:CC2D-SSL}, we first extract template patches for landmarks defined in the annotated template image. Then we use $E_p, E_r$ to embed the patches and query image, resulting in pixel-wise template features and query features. As well as the training stage, we extract the anchor features $F_a$ according to the corresponding coordinate (the red point in Fig.~\ref{fig:CC2D-SSL}). Next, we compute cascade cosine similarity maps $s^i$ for all layers and clip to range $[0, 1]$. At last, we multiply the similarity maps $s^i$ and generate the final prediction by \emph{argmax} operator. In CC2D, we perform inference on the training set to generate pseudo-labels.

\textbf{Feature extractor:} We set the encoder of U-Net~\cite{ronneberger2015u} as a VGG19~\cite{simonyan2014very}. Additionally, we use ASPP~\cite{chen2017rethinking} and a convolution layer with a 1x1 kernel after every up-sampling layer to generate cascade feature embeddings. The detailed illustration can be found in the supplemental materials.

\subsection{Stage II: Training with pseudo-labels (CC2D-TPL)}
\label{Sec:Stage_2}

\begin{figure}[t]
\centering
\includegraphics[width=0.9\textwidth]{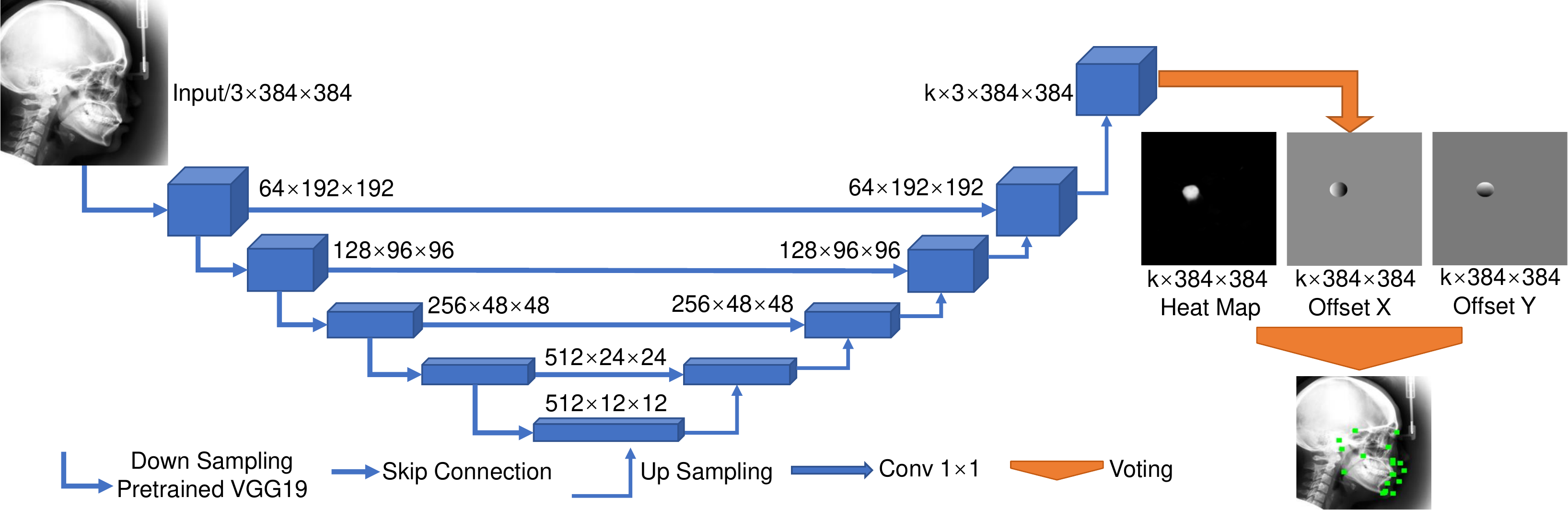}
\caption{The architecture of the multi-task U-Net used in CC2D-TPL. We fine-tune the encoder of U-Net initialized by the VGG19 network \cite{simonyan2014very} pretrained on ImageNet \cite{deng2009imagenet}.} 
\label{fig:CC2D-TPL}
\end{figure}

In this stage, we train a new CNN-based landmark detector from scratch with the pseudo-labels generated in Sec.~\ref{Sec:stage_1}, using the architecture in Fig.~\ref{fig:CC2D-TPL}. We utilize the multi-task U-Net~\cite{yao2020miss} $g$ as our backbone, which predicts both heat map and offset maps simultaneously and has satisfactory performance~\cite{chen2019cephalometric}. Specifically, for the $k^{th}$ landmark located at $(x_k,y_k)$ in an image $X$. We set the ground-truth heat map $Y_{k}^h$ to be 1 if $\sqrt{(x-x_k)^2+(y-y_k)^2} \leq \sigma$ and 0 otherwise. Then, the ground-truth $x$-offset map $Y_{k}^{o_x}$ predicts the relative offset vector $Y_{k}^{o_x} = (x-x_k)/\sigma$ from $x$ to the corresponding landmark $x_k$. Similarly, its $y$-offset map $Y_{k}^{o_y}$ is defined~\cite{yao2020miss}.  The loss function $L_{k}$ for the $k^{th}$ landmark consists of a binary cross-entropy loss, denoted by $L^h$, for punishing the divergence of predicted and ground-truth heatmaps, and the L$_1$ loss, denoted by $L^o$, for punishing the difference in coordinate offset maps.
\begin{equation}
L_{k} = L_{k}^h(Y_{k}^h, g_{k}^h(X)) + sign(Y_{k}^h)\sum_{o \in \{o_x,o_y\}}L_{k}^{o}(Y_{k}^{o}, g_{k}^{o}(X)),
\label{eq:loss_i}
\end{equation}
where $g_{k}^h(X)$ and $g_{k}^o(X)$ are the networks that predict heatmaps and coordinate offset maps, respectively; $sign(\cdot)$ is a sign function which is used to make sure that only the area highlighted by heatmap is included for calculation. At last, we sum the losses $L_k$ for all of the landmarks defined in the template image: 
$L_{TPL} = \sum_k L_k$.
In the test phase, we conduct a majority-vote for candidate landmarks among all pixels with heatmap value $g_{i}^h(X, \theta) \ge  0.5$, according to their coordinate offset maps in $g_{i}^o(X)$. The winning position in the $k^{th}$ channel is the final predicted $k^{th}$ landmark~\cite{chen2019cephalometric}.

\begin{table}[h]
\centering
\caption{Comparison of the state-of-the-art supervised approaches and our CC2D on the ISBI 2015 Challenge~\cite{wang2016benchmark} testset. * represents the performances copied from their original papers. \# represents the performances we re-implement with limited labeled images. We additionally evaluate the performance of CC2D-SSL on the test set.}
\begin{tabular}{|l|c|ccccc|}
\hline
\multirow{2}{*}{Model} & \multirow{2}{*}{\tabincell{l}{Labeled \\ images}} & \multirow{2}{*}{\tabincell{c}{MRE ($\downarrow$) \\ (mm)}} &  \multicolumn{4}{c|}{SDR ($\uparrow$) (\%)} \\ \cline{4-7}
 & &  & 2mm & 2.5mm & 3mm & 4mm \\ \hline
Ibragimov et al.~\cite{ibragimov2015computerized}* & 150 & - & 68.13 & 74.63 & 79.77 & 86.87\\
Lindner et al.~\cite{lindner2015fully}* & 150 & 1.77 & 70.65 & 76.93 & 82.17 & 89.85\\
Urschler et al.~\cite{ref_urschler}* & 150 & - & 70.21 & 76.95 & 82.08 & 89.01\\
Payer et al.~\cite{ref_scn}* & 150 & - & 73.33 & 78.76 & 83.24 & 89.75\\
\hline
Payer et al.~\cite{ref_scn}\# & 25 & 2.54 & 66.12 & 73.27 & 79.82 & 86.82\\
Payer et al.~\cite{ref_scn}\# & 10 & 6.52 & 49.49 & 57.91 & 65.87 & 75.07\\
Payer et al.~\cite{ref_scn}\# & 5 & 12.34 & 27.35 & 32.94 & 38.48 & 45.28\\
\hline
CC2D-SSL & 1 & 4.67 & 40.42 & 47.68 & 55.54 & 68.38  \\
CC2D & 1 & 2.72 & 49.81 & 58.73 & 68.18 & 81.01\\
\hline
\end{tabular}
\label{Table:Main}
\end{table}

\begin{figure}[h]
\centering
\includegraphics[width=1\textwidth]{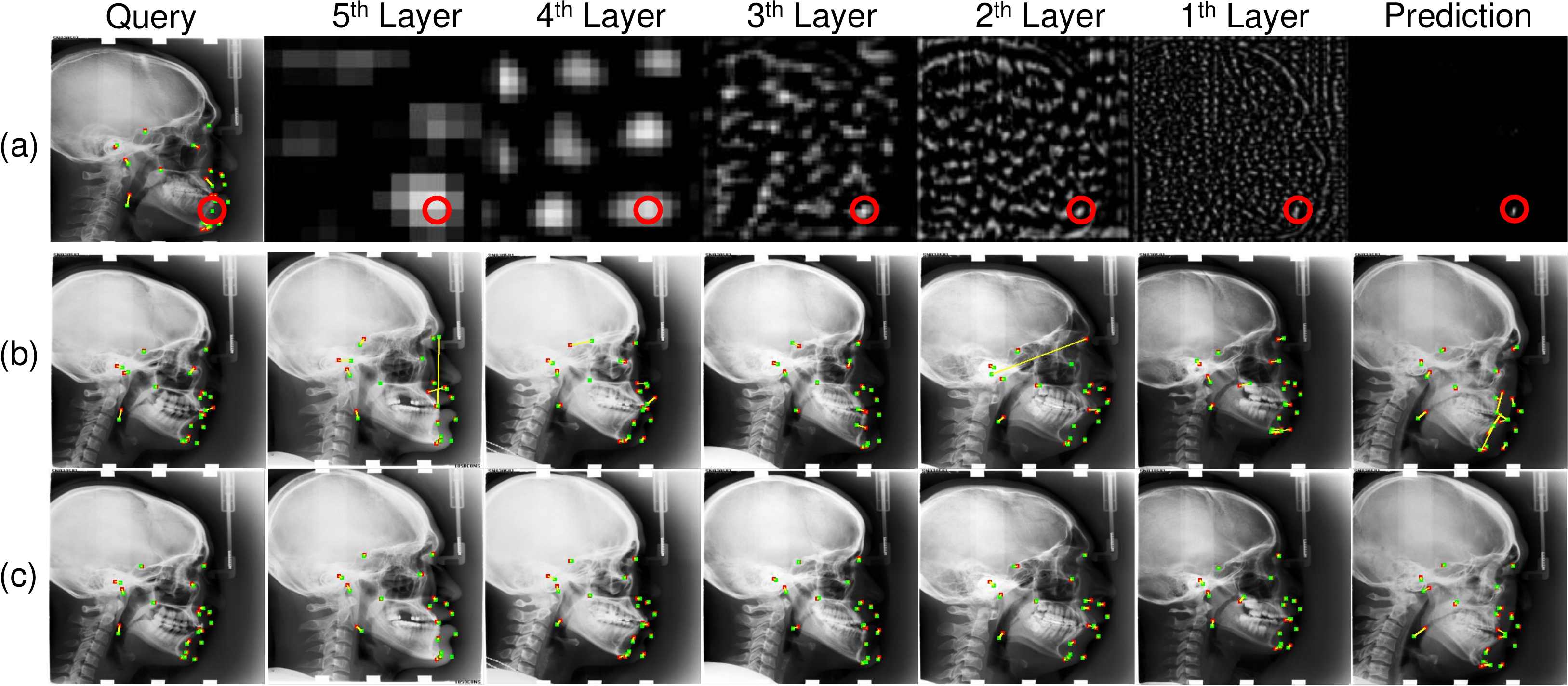}
\caption{(a) The inference procedure of CC2D-SSL for the $6^{th}$ landmark in the query image. The similarity maps at different scales localize the target landmark in a coarse-to-fine fashion. We mark the correct positions by the red circles. (b) Visualizations of the testing images predicted by CC2D-SSL. The landmarks in green and red represent the predictions and ground truths, while the yellow lines mark their distances. (c) Visualizations of the testing images predicted by CC2D. The rightmost images in (b) and (c) are the failure cases. } 
\label{fig:Results}
\end{figure}

\section{Experiments}

\subsection{Settings}
\bheading{Dataset:}
This study uses a widely-used public dataset for cephalometric landmark detection containing 400 radiographs, provided in IEEE ISBI 2015 Challenge~\cite{wang2016benchmark,link_dataset}. Two expert doctors labeled 19 landmarks of clinical anatomical significance for each radiograph. We compute the average annotations by two doctors as the ground truth. The image size is $1935 \times 2400$ and the pixel spacing is 0.1mm. According to the official website, the dataset is split to a training and a test subsets with 150 and 250 radiographs, respectively.

\bheading{Metrics:} Following the official challenge, we use mean radial error (MRE) to measure the Euclidean distance between prediction and ground truth, and successful detection rate (SDR) in four radii (2mm, 2.5mm, 3mm, and 4mm).
%

\bheading{Implementation details:}
All of our models are implemented in PyTorch, accelerated by an NVIDIA RTX3090 GPU.
For self-supervised training, the two feature extractors are optimized by Adam optimizer for 3500 epochs with a learning rate of 0.001 decayed by half every 500 epochs, which takes about 6 hours to converge with a batch size of 8. The embedding length $L$ is set to 16 and the length $\alpha$ of $m$ is 19. For training with pseudo-labels, the multi-task U-Net is optimized by Adam optimizer for 300 epochs with a learning rate of 0.0003, which takes about 50 minutes to converge with a batch size of 16.

\subsection{The performance of CC2D}

During the inference stage of CC2D-SSL, the similarity map in the deepest layer detects the correct coarse area first, then the similarity maps in the inner layers gradually improve the localization accuracy. The procedure is illustrated in Fig.~\ref{fig:Results}(a). Consequently, most of the landmarks in Fig.~\ref{fig:CC2D-SSL}(b) are successfully detected. Moreover, after training by the predictions of CC2D-SSL on the training set, the new landmark detector in CC2D-TPL localizes the landmarks with better accuracy (as shown in Fig.~\ref{fig:Results}(c) and Table~\ref{Table:Main}). 

We quantitatively compare our CC2D with the first~\cite{lindner2015fully} and second~\cite{ibragimov2015computerized} place in the ISBI 2015 Challenge~\cite{wang2016benchmark} in Table~\ref{Table:Main}, as well as two recent state-of-the-art supervised method~\cite{ref_scn,ref_urschler}. With one labeled image available, CC2D achieves the MRE of 2.72 mm and 4mm SDR of 81.01\%, which are competitive compared to the supervised methods. Furthermore, when the available annotated data is reduced, our CC2D shows more superiority. As reported in Table~\ref{Table:Main}, our CC2D performs better than Payer et al.~\cite{payer2016regressing} retrained with 10 labeled images. However, CC2D localizes landmarks with more deviation if there is a drastic difference between the query (failure cases in Fig.~\ref{fig:Results}) and template image (in Fig.~\ref{fig:CC2D}).

\subsection{Ablation study and hyper-parameter analysis}

\textbf{Hyper-parameter analysis: } We study the influence of the embedding length $L$ and the length $\alpha$ of the matrix of interests. According to Table \ref{Table:Hyper}, all of the experimental results fluctuate slightly when changing the two parameters, while setting $L=16$ and $\alpha=19$ leads to the best performance. For $L$, features with larger length may tend to overfit while smaller length may not represent the anatomical information effectively. For $\alpha$, too small matrix has limited receptive field. On the contrary, too large matrix involves too many negative pixels, making the convergence of $L_{SSL}$ (in Eq.~\ref{Eq:L_SSL}) difficult. 

\textbf{Ablation study:} We compare the CC2D using similarity maps in different layers. As shown in (Table \ref{Table:Layer}), the similarity map in the fifth (deepest) layer is most important which selects the coarsest and accurate area first, while other similarity maps also contribute to the final detection accuracy.

\begin{table}[t]
\centering
\caption{The performances of CC2D with different embedding lengths $L$ and different $\alpha$, which is the length of the matrix of interests $m$.}
\begin{tabular}{|c|c|ccccc|}
\hline
\multirow{2}{*}{Para.} & \multirow{2}{*}{Value} & \multirow{2}{*}{\tabincell{c}{MRE ($\downarrow$) \\ (mm)}} &  \multicolumn{4}{c|}{SDR ($\uparrow$) (\%)} \\ \cline{4-7}
 & &  & 2mm & 2.5mm & 3mm & 4mm \\ \hline
\multirow{5}{*}{$L$} & 128 & 3.54 & 37.03 & 46.06 & 56.04 & 71.13\\
& 64 & 3.22 & 40.35 & 49.32 & 60.04 & 73.34\\
& 32 & 2.96 & 45.01 & 54.33 & 54.14 & 78.06\\
& \textbf{16} & \textbf{2.72} & \textbf{49.81} & \textbf{58.73} & \textbf{68.18} & \textbf{81.01}\\
& 8 & 3.22 & 42.23 & 51.20 & 60.37 & 73.66\\
\hline
\multirow{7}{*}{$\alpha$} & 9 & 3.31 & 39.49 & 48.48 & 58.10 & 72.31\\
& 11 & 3.28 & 39.55 & 49.01 & 59.38 & 72.27\\
& 13 & 3.01 & 44.90 & 54.02 & 62.77 & 76.44\\
 & 15 & 2.92 & 47.55 & 56.35 & 65.97 & 78.54\\
 & 17 & 2.85 & 47.83 & 56.75 & 66.21 & 78.37\\
 & \textbf{19} & \textbf{2.72} & \textbf{49.81} & \textbf{58.73} & \textbf{68.18} & \textbf{81.01}\\
& 21 & 2.92 & 47.47 & 56.50 & 65.70 & 77.11\\
\hline
\end{tabular}
\label{Table:Hyper}
\end{table}

\begin{table}[t]
\centering
\caption{The performances of CC2D using cosine similarity maps in different layers.}
\begin{tabular}{|ccccc|ccccc|}
\hline
\multicolumn{5}{|c|}{Layer Index} & \multirow{2}{*}{\tabincell{c}{MRE ($\downarrow$) \\ (mm)}} &  \multicolumn{4}{c|}{SDR ($\uparrow$) (\%)} \\\cline{1-5}\cline{7-10}
5 & 4 & 3 & 2 & 1 & & 2mm & 2.5mm & 3mm & 4mm \\ \hline
$\times$ & $\checkmark$ & $\checkmark$ & $\checkmark$ & $\checkmark$ & 57.96 & 10.07 & 12.45 & 15.01 & 17.89\\
$\checkmark$ & $\times$ & $\checkmark$ & $\checkmark$ & $\checkmark$ & 3.67 & 37.37 & 46.90 & 56.04 & 68.73\\
$\checkmark$ & $\checkmark$ & $\times$ & $\checkmark$ & $\checkmark$ & 10.63 & 23.07 & 27.81 & 31.74 & 37.85\\
$\checkmark$ & $\checkmark$ & $\checkmark$ & $\times$ & $\checkmark$ & 4.67 & 25.72 & 32.75 & 41.87 & 56.27\\
$\checkmark$ & $\checkmark$ & $\checkmark$ & $\checkmark$ & $\times$ & 4.89 & 22.73 & 29.43 & 38.48 & 54.21\\
$\checkmark$ & $\checkmark$ & $\checkmark$ & $\checkmark$ & $\checkmark$ & \textbf{2.72} & \textbf{49.81} & \textbf{58.73} & \textbf{68.18} & \textbf{81.01}\\
\hline
\end{tabular}
\label{Table:Layer}
\end{table}

\section{Conclusion and Future work}

In this paper, we propose Cascade Comparing to Detect (CC2D), a novel framework for building a robust landmark detection network with only one labeled image available. CC2D learns to map the consistent anatomical information into cascading feature spaces by solving a self-supervised patch matching problem. Using the self-supervised model, CC2D localizes the target landmark according to the one-shot template image in a coarse-to-fine fashion to generate pseudo-labels for training a final landmark detector. Extensive experiments evaluate the competitive performance of CC2D, comparable to the state-of-the-art fully-supervised methods. In future work, we plan to further improve the detection accuracy by considering the usage of the spatial relationships of different landmarks.


\section{Supplemental Material}
\begin{figure}
\centering
\includegraphics[width=0.75\textwidth]{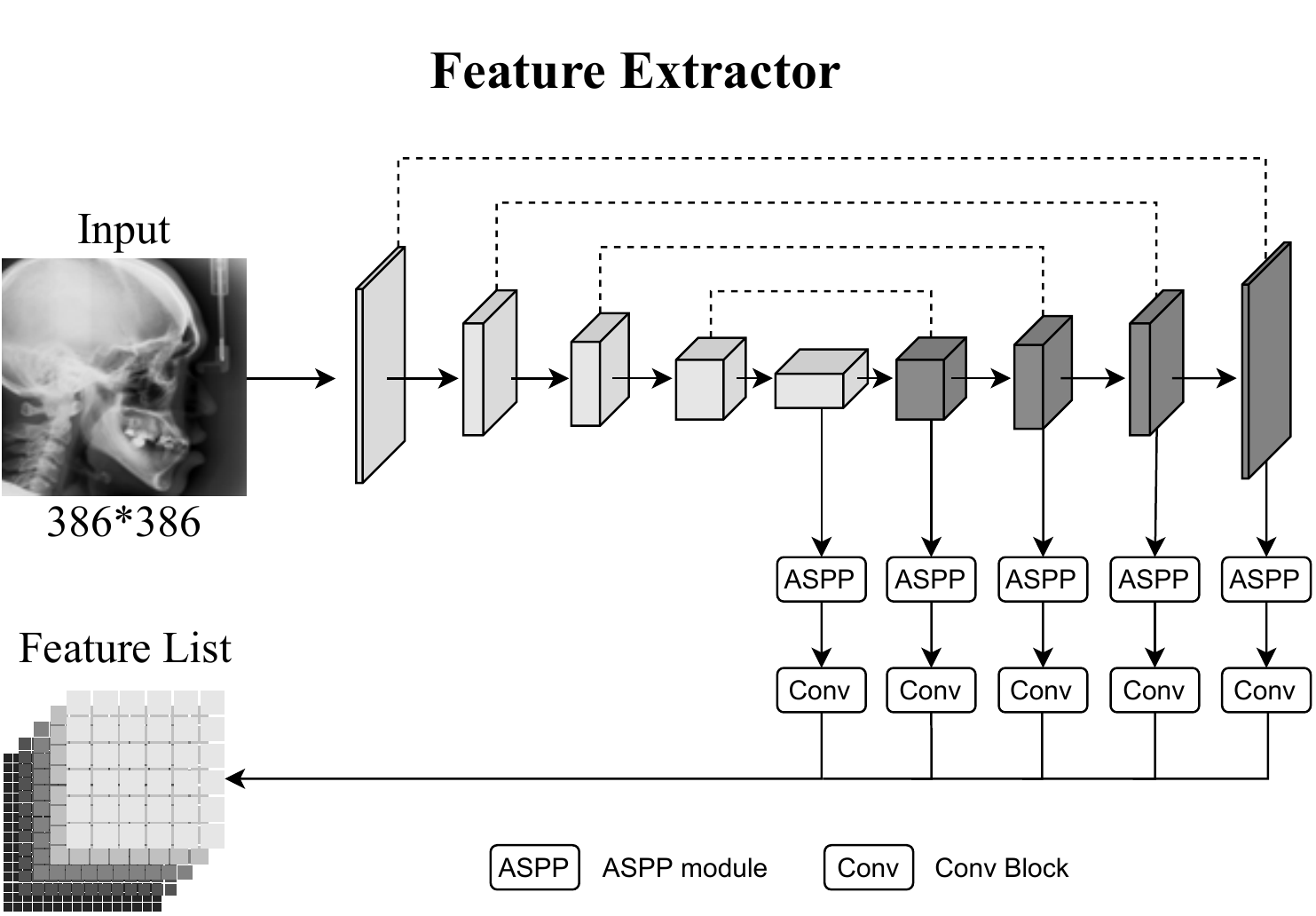}
\caption{The architecture of Feature Extractor used in CC2D-SSL. The \textbf{light grey} part is VGG-style encoder, and \textbf{dark grey} part is decoder. Each dark grey block takes both the output of a block in encoder (dotted lines) and the prior dark grey block (black arrows) as input. The embeddings from dark grey block and last light grey block are carried to ASPP module~\cite{chen2017rethinking} and a Conv block to output the final features.} 
\label{fig:feature_extractor}
\end{figure}

\end{document}


\title{One-Shot Medical Landmark Detection \\
\textit{Supplementary Material}}
%
%
\author{Anonymous}
\authorrunning{Anonymous}
\maketitle       

\begin{figure}
\centering
\includegraphics[width=0.75\textwidth]{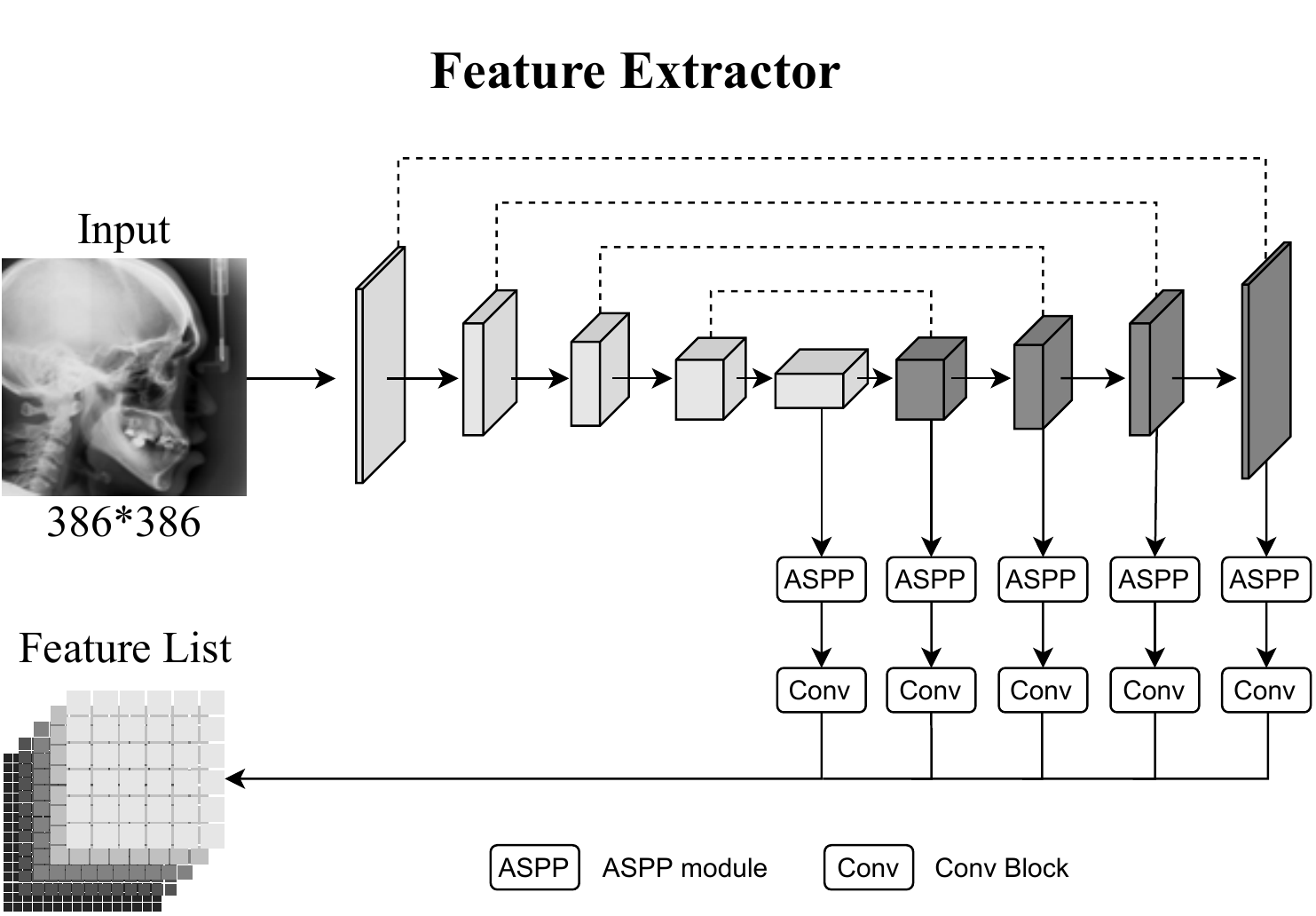}
\caption{The architecture of Feature Extractor used in CC2D-SSL. The \textbf{light grey} part is VGG-style encoder, and \textbf{dark grey} part is decoder. Each dark grey block takes both the output of a block in encoder (dotted lines) and the prior dark grey block (black arrows) as input. The embeddings from dark grey block and last light grey block are carried to ASPP module~\cite{chen2017rethinking} and a Conv block to output the final features.} 
\label{fig:feature_extractor}
\end{figure}



\bibliographystyle{splncs04}
\bibliography{Main}